\documentclass{article}
\usepackage[final]{nips_2016}

\usepackage[utf8]{inputenc} 
\usepackage[T1]{fontenc}    
\usepackage{hyperref}       
\usepackage{url}            
\usepackage{booktabs}       
\usepackage{amsfonts}       
\usepackage{nicefrac}       
\usepackage{microtype}      

\usepackage{natbib}
\usepackage{graphicx}
\graphicspath{ {./graphs/}}
\usepackage{subcaption}
\usepackage{amsmath}
\usepackage{multirow}

\title{Increasing the Interpretability of Recurrent Neural Networks Using Hidden Markov Models}

\author{
  Viktoriya Krakovna \\
  Department of Statistics\\
  Harvard University\\
  \texttt{vkrakovna@fas.harvard.edu} \\
  \And
  Finale Doshi-Velez \\
  Department of Computer Science \\
  Harvard University\\
  \texttt{finale@seas.harvard.edu} \\
}

\begin{document} 

\maketitle

\section{Introduction}
Following the recent progress in deep learning, researchers and practitioners of machine learning are recognizing the importance of understanding and interpreting what goes on inside these black box models. Recurrent neural networks have recently revolutionized speech recognition and translation, and these powerful models would be very useful in other applications involving sequential data. However, adoption has been slow in applications such as health care, where practitioners are reluctant to let an opaque system make crucial decisions. If we can make the inner workings of RNNs more interpretable, more applications can benefit from their power.

A model or algorithm can be considered intelligible to humans in multiple ways, falling under the broad categories of transparency and post-hoc interpretability \citep{Lipton}. While works such as \cite{Ribeiro} and \cite{Turner} develop post-hoc explanations for black-box models, we focus on transparency, specifically model parsimony and the ability to trace back from a prediction or model component to particularly influential features in the data, similarly to \cite{Kim}.

This could be useful for understanding mistakes made by neural networks, which have human-level performance most of the time, but can perform very poorly on seemingly easy cases. For instance, convolutional networks can misclassify adversarial examples with very high confidence \citep{Szegedy}, and made headlines in 2015 when the image tagging algorithm in Google Photos mislabeled African Americans as gorillas. It's reasonable to expect recurrent networks to fail in similar ways as well. It would thus be useful to have more visibility into where these sorts of errors come from, i.e. which groups of features contribute to such flawed predictions. 

Several promising approaches to interpreting RNNs have been developed recently, focusing on a state-of-the-art RNN architecture called Long Short-Term Memory (LSTM). \citet{Che} use gradient boosting trees to predict LSTM output probabilities and explain which features played a part in the prediction. They do not model the internal structure of the LSTM, but instead approximate the entire architecture as a black box. \citet{Karpathy} showed that in LSTM language models, around 10\% of the memory state dimensions can be interpreted with the naked eye by color-coding the text data with the state values; some of them track quotes, brackets and other clearly identifiable aspects of the text. Building on these results, we take a somewhat more systematic approach to looking for interpretable hidden state dimensions, by using decision trees to predict individual hidden state dimensions (Figure \ref{fig:decision_trees}). We visualize the overall dynamics of the LSTM hidden states by coloring the training data with the k-means clusters on the state vectors (Figure \ref{subfig:linux_lstm}).

We explore several methods for building more interpretable models by combining LSTMs and HMMs. The existing body of literature mostly focuses on methods that specifically train the RNN to predict HMM states \citep{Bourlard} or posteriors \citep{Maas}, referred to as hybrid or tandem methods respectively.
We add the HMM state probabilities to the output layer of the LSTM, and then train the HMM and LSTM either sequentially or jointly (Figure \ref{fig:models}). The LSTM model can make use of the information from the HMM, and fill in the gaps when the HMM is not performing well. This results in an LSTM with a smaller number of hidden state dimensions that could be interpreted individually, especially for smaller data sets. We test the algorithms on text data and multivariate medical data.

\section{Methods}

\subsection{LSTM} 

We use a character-level LSTM with 1 layer and no dropout, based on the Element-Research library. We train the LSTM for 10 epochs, starting with a learning rate of 1, where the learning rate is halved whenever $\exp(-l_t) > \exp(-l_{t-1}) + 1$, where $l_t$ is the log likelihood score at epoch $t$. The $L_2$-norm of the parameter gradient vector is clipped at a threshold of 5.


\begin{figure}[t]
\begin{subfigure}{0.5\textwidth}
\includegraphics[scale=.2]{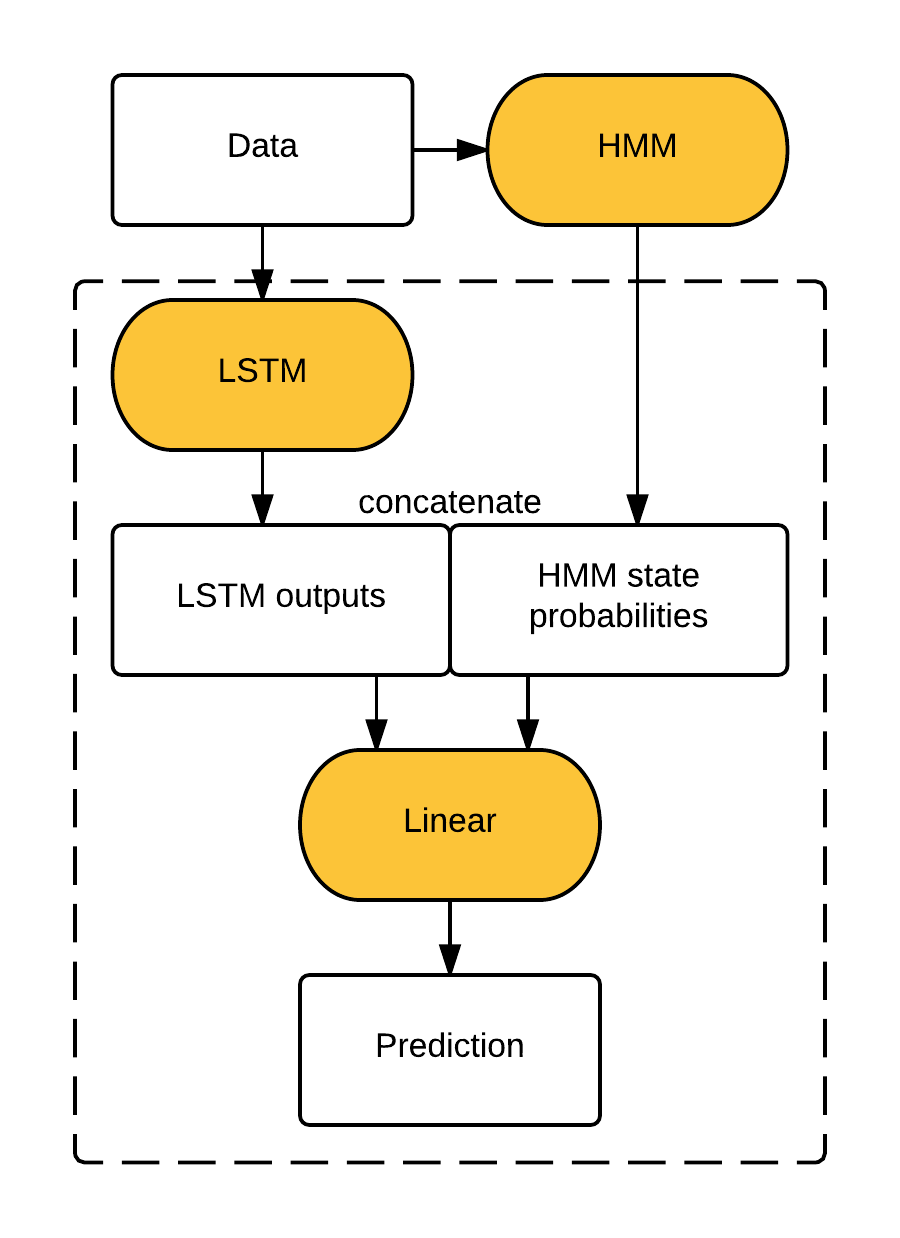}
\caption{Sequentially trained hybrid algorithm}
\label{subfig:seq_hybrid}
\end{subfigure}%
\hfill
\begin{subfigure}{0.5\textwidth}
\includegraphics[scale=.2]{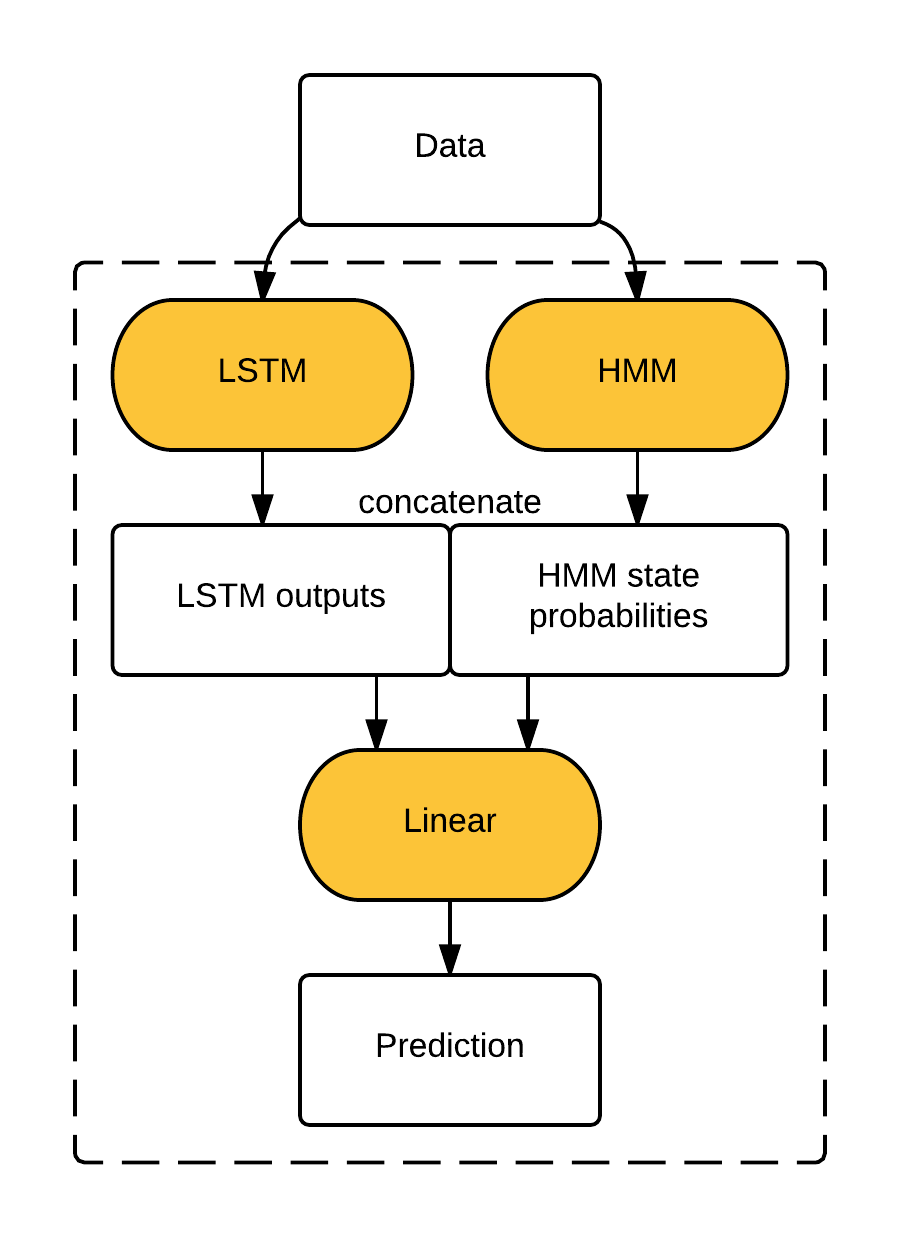}
\caption{Jointly trained hybrid algorithm}
\label{subfig:joint_hybrid}
\end{subfigure}
\caption{Hybrid HMM-LSTM algorithms (the dashed blocks indicate the components trained using SGD in Torch).}
\label{fig:models}
\end{figure}

\subsection{Hidden Markov models}



We initialization the HMM hidden states as a random multinomial draw for each time step (i.i.d. across time steps). Then at each iteration:

\begin{enumerate}
	\item Sample states using Forward Filtering Backwards Sampling algorithm (FFBS, \citet{Rao}).
	\item Sample transition parameters from a Multinomial-Dirichlet posterior with hyperparameter $\alpha$.
	Let $n_{ij}$ be the number of transitions from state $i$ to state $j$. Then the posterior distribution of the $i$-th row of transition matrix $T$ (corresponding to transitions from state $i$) is: 
	$$T_i \sim \text{Mult}(n_{ij} | T_i) \text{Dir}(T_i | \alpha)$$
    
    \item Sample the emission parameters from a Multinomial-Dirichlet posterior.
\end{enumerate}

%

\subsection{Hybrid models}

Our main hybrid model is put together sequentially, as shown in Figure \ref{subfig:seq_hybrid}. We first run the discrete HMM on the data, outputting the hidden state distributions obtained by the HMM's forward pass, and then add this information to the architecture in parallel with a 1-layer LSTM. The linear layer between the LSTM and the prediction layer is augmented with an extra column for each HMM state. The LSTM component of this architecture can be smaller than a standalone LSTM, since it only needs to fill in the gaps in the HMM's predictions. The HMM is written in Python, and the rest of the architecture is in Torch.

We also build a joint hybrid model, where the LSTM and HMM are simultaneously trained in Torch, as shown in Figure \ref{subfig:joint_hybrid}. We implemented an HMM Torch module, optimized using stochastic gradient descent rather than FFBS. Similarly to the sequential hybrid model, we concatenate the LSTM outputs with the HMM state probabilities.  


\begin{table}[t]
\caption{Predictive loglikelihood (LL) comparison, sorted by validation set performance.}
\label{tab:results}
\begin{subtable}{.5\linewidth}
\centering
\caption{Linux text data}\label{tab:linux}
\begin{tabular}{l|p{1cm}|p{1cm}|l}
       Method         & LSTM \newline dim & HMM \newline states & LL  \\\hline
       HMM            &     & 10 & -2.76 \\
       HMM            &     & 20 & -2.55 \\
       LSTM           & 5   &    & -2.54 \\
       Joint hybrid   & 5   & 10 & -2.37 \\
       Hybrid         & 5   & 10 & -2.33 \\
       Hybrid         & 5   & 20 & -2.25 \\
       Joint hybrid   & 10  & 10 & -2.19 \\
       LSTM           & 10  &    & -2.17 \\
       Hybrid         & 10  & 10 & -2.14 \\
       Hybrid         & 10  & 20 & -2.07 \\
       LSTM           & 15  &    & -2.03 \\
       Joint hybrid   & 15  & 10 & -2.00 \\
       Hybrid         & 15  & 10 & -1.96 \\
       Hybrid         & 15  & 20 & -1.96 \\
       Joint hybrid   & 20  & 10 & -1.91 \\
       LSTM           & 20  &    & -1.88 \\
       Hybrid         & 20  & 10 & -1.87 \\
       Hybrid         & 20  & 20 & -1.85 \\\hline
\end{tabular}
\end{subtable}\hfill
\begin{subtable}{.5\linewidth}
\centering
\caption{Shakespeare text data}\label{tab:shak}
\begin{tabular}{l|p{1cm}|p{1cm}|l}
       Method         & LSTM \newline dim & HMM \newline states & LL  \\\hline
       HMM            &     & 10 & -2.69 \\
       HMM            &     & 20 & -2.5  \\
       LSTM           & 5   &    & -2.41 \\
       Hybrid         & 5   & 10 & -2.3  \\
       Hybrid         & 5   & 20 & -2.26 \\
       LSTM           & 10  &    & -2.23 \\
       Joint hybrid   & 5   & 10 & -2.21 \\
       Hybrid         & 10  & 10 & -2.19 \\
       Hybrid         & 10  & 20 & -2.16 \\
       Joint hybrid   & 10  & 10 & -2.12 \\
       Hybrid         & 15  & 10 & -2.13 \\
       LSTM           & 15  &    & -2.1  \\
       Hybrid         & 15  & 20 & -2.07 \\
       Hybrid         & 20  & 10 & -2.05 \\
       Joint hybrid   & 15  & 10 & -2.03 \\
       LSTM           & 20  &    & -2.03 \\
       Hybrid         & 20  & 20 & -2.02 \\
       Joint hybrid   & 20  & 10 & -1.97 \\
       \hline
\end{tabular}
\end{subtable}
\begin{subtable}{.5\linewidth}
\centering
\caption{Single-patient Physionet data}\label{tab:single_patient}
\begin{tabular}{p{.9cm}|l|p{.75cm}|p{.7cm}|l}
Method & Features    & LSTM \newline dim & HMM \newline states & LL  \\\hline
HMM    & Discretized &    & 10 & -0.68 \\
LSTM   & Continuous  & 5  &    & -0.63  \\
LSTM   & Continuous  & 10 &    & -0.63  \\
Hybrid & Discretized & 5  & 10 & -0.39  \\
Hybrid & Discretized & 10 & 10 & -0.39  \\
LSTM   & Discretized & 5  &    & -0.37  \\
HMM    & Continuous  &    & 10 & -0.34  \\
LSTM   & Discretized & 10 &    & -0.33  \\
Hybrid & Continuous  & 5  & 10 & -0.33  \\
Hybrid & Continuous  & 10 & 10 & -0.33  \\\hline
\end{tabular}
\end{subtable}\hfill
\begin{subtable}{.5\linewidth}
\centering
\caption{Multi-patient Physionet data}\label{tab:multi_patient}
\begin{tabular}{p{.9cm}|l|p{.75cm}|p{.7cm}|l}
Method & Features    & LSTM \newline dim & HMM \newline states & LL  \\\hline
Hybrid & Discretized & 10 & 10 & -0.61 \\
HMM    & Discretized &    & 10 & -0.60  \\
Hybrid & Discretized & 5  & 10 & -0.59 \\
LSTM   & Discretized & 10 &    & -0.58 \\
LSTM   & Discretized & 5  &    & -0.56 \\
HMM    & Continuous  &    & 10 & -0.54 \\
Hybrid & Continuous  & 5  & 10 & -0.54 \\
Hybrid & Continuous  & 10 & 10 & -0.54 \\
LSTM   & Continuous  & 5  & 10 & -0.54 \\
LSTM   & Continuous  & 10 & 10 & -0.54 \\\hline
\end{tabular}
\end{subtable}
\end{table}

\begin{figure}
\begin{subfigure}{0.49\textwidth}
\centering
\includegraphics[scale=.63]{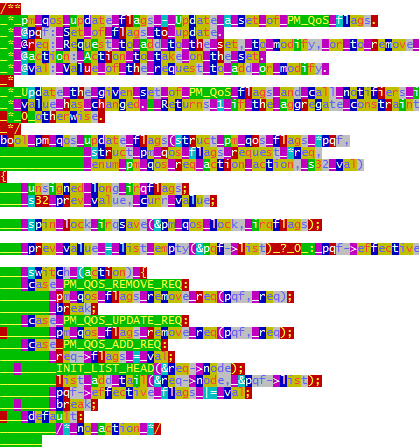}
\caption{Hybrid HMM component: colors correspond to 10 HMM states. Distinguishes comments and indentation spaces (green, yellow font) from other spaces (purple). Red cluster (with yellow font) identifies punctuation and brackets. Green cluster (yellow font) also finds capitalized variable names.}
\label{subfig:linux_hmm}
\end{subfigure}\hfill
\begin{subfigure}{0.49\textwidth}
\centering
\includegraphics[scale=.63]{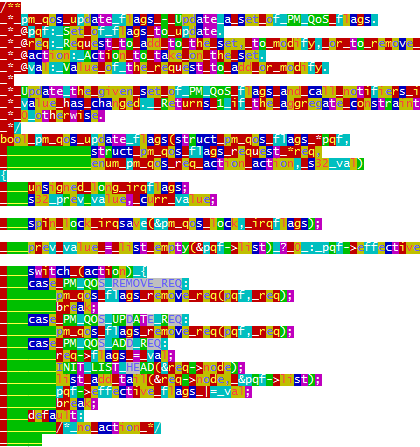}
\caption{Hybrid LSTM component: colors correspond to 10 k-means clusters on hidden state vectors. Distinguishes comments, spaces at beginnings of lines, and spaces between words (red, white font) from indentation spaces (green, yellow font). Opening brackets are red (yellow font), closing brackets are green (white font).} \label{subfig:linux_lstm}
\end{subfigure}
\caption{Visualizing HMM and LSTM states on Linux data for the hybrid with 10 LSTM state dimensions and 10 HMM states. The HMM and LSTM components learn some complementary features in the text related to spaces and comments. \label{fig:viz_hybrid_linux}}
\end{figure}

\begin{figure}
\centering
\includegraphics[scale=.45]{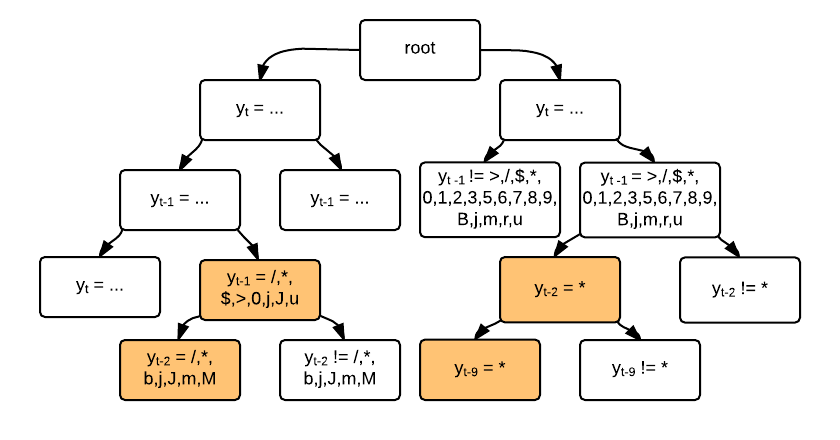}
\caption{Decision tree predicting an individual hidden state dimension of the hybrid algorithm based on the preceding characters on the Linux data. Nodes with uninformative splits are represented with $\dots$.}
\label{fig:decision_trees}
\end{figure}

\section{Experiments}
We test the methods on two text data sets used by \citet{Karpathy}, Tiny Shakespeare (1M characters) and Linux Kernel (5M characters). We also use ICU medical data from the 2014 Physionet challenge, whose objective is to detect heart beat windows using 6 physiological signal features such as ECG measurements. We test on a single-patient data set with 200K time points, predicting future heart beats for the same patient, and a multi-patient data set with 15 training and 5 test patients.

For the text data sets, Tables \ref{tab:linux} and \ref{tab:shak} shows the predictive log likelihood for the next text character for each method. The sequential hybrid algorithm performs a bit better than the standalone LSTM with the same LSTM state dimension. This effect gets smaller as we increase the LSTM size and the HMM makes less difference to the prediction. The hybrid algorithm with 20 HMM states does better than the one with 10 HMM states. The joint hybrid algorithm outperforms the sequential hybrid on Shakespeare data, but does worse on Linux data, which suggests that the joint hybrid is more helpful for smaller data sets. The joint hybrid is an order of magnitude slower than the sequential hybrid, as the SGD-based HMM is slower to train than the FFBS-based HMM.


We interpret the HMM and LSTM states in the hybrid algorithm with 10 LSTM state dimensions and 10 HMM states in Figure \ref{fig:viz_hybrid_linux}, showing which text features are identified by the HMM and LSTM components for the Linux data. In Figure \ref{subfig:linux_hmm}, we color-code the training data with the 10 HMM states. In Figure \ref{subfig:linux_lstm}, we apply k-means clustering to the LSTM state vectors, and color-code the training data with the clusters. The HMM and LSTM states pick up on spaces, indentation, and special characters (such as comment symbols in Linux data). Sometimes, the HMM and LSTM complement each other, such as learning different features related to spaces and comments in the Linux data. 
In Figure \ref{fig:decision_trees}, we see that some individual LSTM hidden state dimensions identify similar features, such as comment symbols in the Linux data. The 10 hidden state dimensions of the hybrid algorithm mostly track comment characters, which suggests these features have a distributed representation.

For the Physionet data, we try the methods on continuous features and discretized features. Tables \ref{tab:single_patient} and \ref{tab:multi_patient} shows the predictive log likelihood for the heart beat indicator for each method. Curiously, discretizing the single-patient data makes the LSTM perform better, while the HMM and hybrid perform worse, with the hybrid on continuous features and LSTM of dimension 10 on discretized features performing the best. We do not observe this effect on the multi-patient data set, where all the methods perform better (and similarly) on continuous features. The LSTM dimension mostly doesn't matter for these data sets. The hybrid algorithm does not decrease the state dimension for the multi-patient data set, while the smaller single-patient data the hybrid of dimension 5 performs the same as the LSTM of dimension 10 without relying on discretization.

\section{Conclusion}
Hybrid HMM-RNN approaches combine the interpretability of HMMs with the predictive power of RNNs. A small hybrid model usually performs better than a standalone LSTM of the same size, especially on smaller data sets. We use visualizations to show how the LSTM and HMM components of the hybrid algorithm complement each other in terms of features learned in the data.

\bibliography{int_bibliography}

\begin{thebibliography}{}

\bibitem[Bourlard and Morgan, 1994]{Bourlard}
Bourlard, H. and Morgan, N. (1994).
\newblock {\em {Connectionist Speech Recognition: A Hybrid Approach}}.
\newblock Kluwer Academic Publishers.

\bibitem[Che et~al., 2015]{Che}
Che, Z., Purushotham, S., and Liu, Y. (2015).
\newblock {Distilling Knowledge from Deep Networks with Applications to
  Healthcare Domain}.
\newblock {\em {Neural Information Processing Systems Workshop on Machine
  Learning for Healthcare (MLHC)}}.

\bibitem[Karpathy et~al., 2016]{Karpathy}
Karpathy, A., Johnson, J., and Fei-Fei, L. (2016).
\newblock {Visualizing and Understanding Recurrent Networks}.
\newblock {\em International Conference for Learning Representations Workshop
  Track}.

\bibitem[Kim et~al., 2015]{Kim}
Kim, B., Shah, J.~A., and Doshi-Velez, F. (2015).
\newblock {Mind the Gap: A Generative Approach to Interpretable Feature
  Selection and Extraction}.
\newblock In Cortes, C., Lawrence, N.~D., Lee, D.~D., Sugiyama, M., and
  Garnett, R., editors, {\em Neural Information Processing Systems (NIPS)},
  pages 2260--2268.

\bibitem[Lipton, 2016]{Lipton}
Lipton, Z.~C. (2016).
\newblock {The Mythos of Model Interpretability}.
\newblock {\em 2016 ICML Workshop on Human Interpretability in Machine Learning
  (WHI), New York, NY}.

\bibitem[Maas et~al., 2012]{Maas}
Maas, A., Le, Q., O'Neil, T., Vinyals, O., Nguyen, P., and Ng, A. (2012).
\newblock {Recurrent Neural Networks for Noise Reduction in Robust ASR}.
\newblock In {\em Proceedings of INTERSPEECH}.

\bibitem[Rao and Teh, 2013]{Rao}
Rao, V. and Teh, Y.~W. (2013).
\newblock Fast {MCMC} sampling for {M}arkov jump processes and extensions.
\newblock {\em Journal of Machine Learning Research}, 14:3207--3232.
\newblock arXiv:1208.4818.

\bibitem[Ribeiro et~al., 2016]{Ribeiro}
Ribeiro, M.~T., Singh, S., and Guestrin, C. (2016).
\newblock {"Why Should {I} Trust You?": Explaining the Predictions of Any
  Classifier}.
\newblock {\em CoRR}, abs/1602.04938.

\bibitem[Szegedy et~al., 2014]{Szegedy}
Szegedy, C., Zaremba, W., Sutskever, I., Bruna, J., Erhan, D., Goodfellow, I.,
  and Fergus, R. (2014).
\newblock Intriguing properties of neural networks.
\newblock In {\em International Conference on Learning Representations}.

\bibitem[Turner, 2016]{Turner}
Turner, R. (2016).
\newblock {A Model Explanation System: Latest Updates and Extensions}.
\newblock {\em 2016 ICML Workshop on Human Interpretability in Machine Learning
  (WHI), New York, NY}.

\end{thebibliography}
\bibliographystyle{apalike}

\end{document}